\documentclass[conference]{IEEEtran}
\IEEEoverridecommandlockouts
\usepackage{cite}
\usepackage{amsmath,amssymb,amsfonts}
\usepackage{algorithmic}
\usepackage{graphicx}
\usepackage{textcomp}
\usepackage{xcolor}
\usepackage{orcidlink}
\usepackage{fancyhdr}
\hypersetup{
    hidelinks
}

\fancypagestyle{firstpageheader}{
    \fancyhf{} 
    \fancyhead[C]{\footnotesize PREPRINT: Copyright IEEE 2024. This is the author's version of the work. It is posted here for your personal use. Not for redistribution. The definitive version is published in The 11th IEEE International Conference on Software Defined Systems (SDS-2024).}
}

\makeatletter
\newcommand{\linebreakand}{%
  \end{@IEEEauthorhalign}
  \hfill\mbox{}\par
  \mbox{}\hfill\begin{@IEEEauthorhalign}
}
\makeatother

\usepackage{listings}
\definecolor{codegreen}{rgb}{0,0.6,0}
\definecolor{codegray}{rgb}{0.5,0.5,0.5}
\definecolor{codepurple}{rgb}{0.58,0,0.82}
\definecolor{backcolour}{rgb}{0.95,0.95,0.92}
\definecolor{bluecolour}{rgb}{0,0,0.92}
\lstdefinestyle{mystyle}{
    backgroundcolor=\color{backcolour},   
    commentstyle=\color{codegreen},
    keywordstyle=\color{magenta},
    numberstyle=\tiny\color{codegray},
    stringstyle=\color{codepurple},
    basicstyle=\ttfamily\footnotesize,
    breakatwhitespace=false,         
    breaklines=true,                 
    captionpos=b,                    
    keepspaces=true,                 
    numbers=left,                    
    numbersep=5pt,                  
    showspaces=false,                
    showstringspaces=false,
    showtabs=false,                  
    tabsize=2
}
\begin{document}

\title{Holon Programming Model: A Software-Defined Approach for System of Systems}

\author{
\IEEEauthorblockN{Muhammad Ashfaq\,\orcidlink{0000-0003-1870-7680}}
\IEEEauthorblockA{\textit{Faculty of IT} \\
\textit{University of Jyv\"{a}skyl\"{a}}\\
Jyv\"{a}skyl\"{a}, Finland \\
muhammad.m.ashfaq@jyu.fi}
\and
\IEEEauthorblockN{Ahmed R. Sadik\,\orcidlink{0000-0001-8291-2211}}
\IEEEauthorblockA{
\textit{Senior Scientist}\\
\textit{Honda Research Institute Europe} \\
Offenbach am Main, Germany \\
ahmed.sadik@honda-ri.de}
\and
\IEEEauthorblockN{Tommi Mikkonen\,\orcidlink{0000-0002-8540-9918}}
\IEEEauthorblockA{\textit{Faculty of IT} \\
\textit{University of Jyv\"{a}skyl\"{a}}\\
Jyv\"{a}skyl\"{a}, Finland \\
tommi.j.mikkonen@jyu.fi}
\linebreakand
\IEEEauthorblockN{Muhammad Waseem\,\orcidlink{0000-0001-7488-2577}}
\IEEEauthorblockA{\textit{Faculty of IT} \\
\textit{University of Jyv\"{a}skyl\"{a}}\\
Jyv\"{a}skyl\"{a}, Finland \\
muhammad.m.waseem@jyu.fi}
\and
\IEEEauthorblockN{Niko M\"{a}kitalo\,\orcidlink{0000-0002-7994-3700}}
\IEEEauthorblockA{\textit{Faculty of IT} \\
\textit{University of Jyv\"{a}skyl\"{a}}\\
Jyv\"{a}skyl\"{a}, Finland \\
niko.k.makitalo@jyu.fi}
}

\maketitle
\thispagestyle{firstpageheader} 

\begin{abstract}
As Systems of Systems (SoSs) evolve into increasingly complex networks, harnessing their collective potential becomes paramount. 
Traditional SoS engineering approaches lack the necessary programmability to develop third-party SoS-level behaviours. 
To address this challenge, we propose a \textit{software-defined} approach to enable flexible and adaptive programming of SoS. 
We introduce the \textit{Holon Programming Model (HPM)}, a software-defined framework designed to meet these needs. 
The HPM empowers developers to design and orchestrate complex system behaviours effectively, as illustrated in our disaster management scenario. 
This research outlines the HPM's theoretical underpinnings and practical applications, intending to drive further exploration and advancement in the field of software-defined SoS.

\end{abstract}


\section{Introduction}



A \textit{System of Systems} (SoS) integrates independent systems to achieve higher-level goals, exemplified by smart cities and modern healthcare networks~\cite{nielsen2015systems}. 
As the modern SoS evolves into an increasingly complex and dynamic network, it gives rise to a \textit{dynamic SoS} (dynaSoS), characterized by more frequent contextual changes and adaptive behaviors~\cite{heinrich2023industry}.
This evolution presents new challenges and opportunities in SoS design and management.

The main challenges in a dynaSoS are \textit{interoperability} and \textit{adaptability}~\cite{adler2023research}. Interoperability facilitates seamless communication among heterogeneous constituent systems (CSs) without requiring deep knowledge of their internals. Adaptability enables the SoS to respond dynamically to internal and external changes, allowing CSs to join or leave at runtime.
These challenges become even more complex when considered together, demanding constant reassessment and adjustment of communication protocols and adaptation strategies.



To address these challenges, the \textit{holonic architecture} offers a robust framework for designing and modelling the SoS behaviour. Inspired by the concept of \textit{holons}---autonomous entities that function as both whole systems and parts of larger systems---this architecture enables CSs to operate independently while collaborating to achieve the SoS goals~\cite{blair2015}. Recent advancements in holonic architectures have enabled features like CS interoperability~\cite{nundloll2020, icsoft24}, dynamic composition~\cite{elhabbash2024principled}, and ad-hoc scalability~\cite{sadikSelfadaptiveSystemSystems2023,sadikCombiningAdaptiveHolonic2017}.
However, it needs to improve in effectively programming and coordinating complex, system-wide behaviours in response to dynamic environmental changes~\cite{9529560}.
We hypothesize that these limitations stem from the absence of mechanisms to explicitly program the interactions and behaviours of holons within an SoS, thus hindering their ability to make informed decisions and adapt dynamically.

To address these limitations of the holonic architecture, we propose a software-defined approach, leveraging the principles of \textit{Software-Defined Systems} (SDSs) to enhance the programmability and adaptability of SoSs. SDSs have emerged as a powerful paradigm for managing complex, distributed infrastructures by separating the control plane from the data plane and centralizing system intelligence~\cite{zeng2020software}. Integrating software-defined capabilities into the holonic architecture can overcome current limitations in decision-making and responsiveness, enabling more efficient management and control of complex SoSs~\cite{soare2023software}.

Our proposed \textit{Holon Programming Model} (HPM) is a conceptual framework designed to enhance the programmability of holonic architectures of SoSs. Building upon the principles of software-defined systems, particularly the separation of control logic from the underlying infrastructure, the HPM provides a structured approach for defining and managing the interactions and behaviours of holons. 
As software-defined networking decouples the control plane from the data plane to allow more flexible network management~\cite{kreutz2014software}, the HPM separates the programming of SoS behaviours from the SoS composition and architectural layers. 
This separation provides a structured approach for defining and managing holons' interactions and behaviours, resulting in a more interoperable and adaptive SoS.
In this way, the HPM offers a promising solution to the complex programming needs of modern SoSs.

To demonstrate the practical utility of the HPM, we apply it to a disaster management scenario involving the dynamic composition of heterogeneous entities, such as command centres and government departments, into an SoS. In this scenario, the HPM facilitates the programming and coordination of these entities' behaviours, allowing the SoS to dynamically adapt and effectively manage complex operations, such as search and rescue, in response to evolving disaster conditions.

The key contributions of this paper are:
\begin{itemize}
    \item Introduction of the HPM~(Section~\ref{sec:HPM}), a novel software-defined framework that enhances the programmability and adaptability of SoSs by enabling the secure and dynamic composition of CSs~(Fig.~\ref{fig:holon_composition}) and the orchestration of complex system behaviours through a layered architectural approach (Fig.~\ref{fig:hpm_layered}).
    \item Illustration of HPM's application in a complex, real-world SoS environment through a disaster management scenario (Section~\ref{sec:scenario}).
\end{itemize}

\section{Background and Related Work}
\label{sec:background}

\subsection{System of Systems (SoS)}
An SoS is a collection of independent systems that work together to achieve capabilities and results unattainable by the individual systems alone~\cite{nielsen2015systems}. 
SoSs are characterized by their operational and managerial independence, emergent behaviour, evolutionary development, and geographic distribution~\cite{maier1998}. 
For instance, modern smart energy grids integrate diverse and autonomous systems such as renewable energy sources, energy storage units, and consumer management systems to create a resilient and efficient energy infrastructure~\cite{egert2021holonic}. This integration enables real-time balancing of supply and demand, enhances energy efficiency, and improves the reliability of energy delivery services.

DynaSoSs are a subclass of SoSs that show higher inherent dynamic and adaptability about system change, context, adaptation, and goal modification~\cite{heinrich2023industry}. 
A dynaSoS can be composed of a combination of cyber-physical systems, information systems, autonomous systems, humans, infrastructure, and other potential subsystems. For example, a system of autonomous systems is a dynaSoS formed by integrating autonomous systems, leading to reduced demands for human cognition and workload while enabling new capabilities~\cite{torkjazi2022taxonomy}.

\subsection{Holonic Architecture}

The term `holon' [Greek: \textit{holos} (whole) + \textit{on} (part)] was used to describe the philosophy that every component in a system functions as both a self-sufficient entity \textit{and} a part of a larger whole~\cite{koestlerGhostMachine1968}. In contrast to the atomic view, where elements are regarded as isolated units, his `watchmaker parable' illustrates that complex systems are more stable when built from semi-autonomous subsystems (holons). This concept mirrors how natural systems, from organisms to societies, are organized into nested hierarchies---or \textit{holarchies}---ensuring resilience and flexibility.

\begin{figure}
  \centering
  \includegraphics[width=\columnwidth]{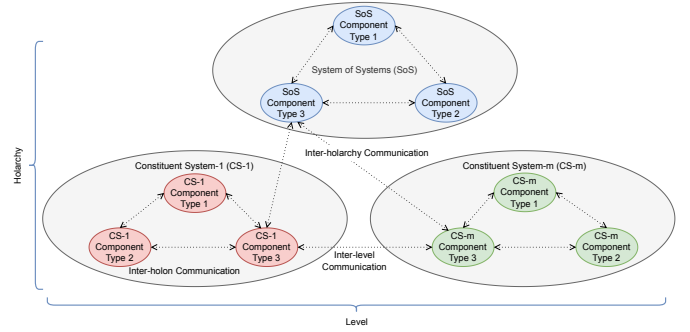}
  \caption{The Holonic Architecture}
  \label{fig:holonic_architecture}
\end{figure}

The holonic architecture pattern was introduced in intelligent manufacturing, where a holon is defined as an autonomous, cooperative building block within a holarchy---a hierarchical organization of holons.
Fig.~\ref{fig:holonic_architecture} shows how holons are organized into levels within a holarchy, offering both vertical and horizontal scalability.
This pattern balances centralized and heterarchical architecture patterns, promoting adaptability and resilience in industrial units.

In the context of SoSs, holonic architecture models each CS as a holon, enabling their programmatic specification, reasoning, and manipulation, which in turn facilitates the composition of new holons (and thus, the construction of an SoS)~\cite{blair2015}.
Recent research has leveraged  \textit{ontologies} where the ontological description of a holon encapsulates the services, capabilities, and properties of a CS~\cite{nundloll2020}.
This empowers CS to comprehend the functionality of other CS, reason about them, and establish effective communication~\cite{elhabbashOntologicalArchitecturePrincipled2020}.

\subsection{Software-defined Systems (SDSs)}

SDSs represent a paradigm shift in system design and management, where the control functions are decoupled from the underlying hardware infrastructure~\cite{zeng2020software}. This separation allows for greater flexibility, programmability, and centralized management of complex systems. The main characteristics of an SDS include hardware resource abstraction, automation, and dynamic reconfiguration, enabling more adaptable and efficient system operations.

While SDSs offer powerful capabilities for managing complex systems, their centralized control can conflict with the fundamental characteristics of SoSs, such as the operational and managerial independence of CSs. Therefore, rather than fully adopting a centralized SDS approach, we selectively apply relevant software-defined principles to enhance the programmability and adaptability of SoSs while maintaining the autonomy and distributed nature of its CSs.

By applying SDS concepts to SoS, we can address critical challenges of interoperability and adaptability. This approach enables real-time adaptation to changing environments and requirements while respecting the autonomy of CSs. Additionally, it facilitates better communication between heterogeneous CSs through standardized interfaces and protocols, similar to how software-defined networking (SDN) addresses the heterogeneity of network devices using abstraction layers and standardized interfaces.

However, applying SDS concepts to SoS requires a specialized programming model tailored to the unique demands of these complex systems. 
Such a model must be reactive to system dynamics and environmental changes, enabling developers to create and manage SoS-level behaviours seamlessly.
Unfortunately, the current SoS landscape lacks a model that fully leverages these principles. This gap necessitates a shift from traditional, sequential, and static development approaches to software-defined ones that manage the collective interactions among CSs more effectively.

\section{The Holon Programming Model}
\label{sec:HPM}


The HPM is designed to efficiently develop coordinated, multi-system programs that are proactively initiated and responsive to environmental changes. By leveraging collective SoS behaviours as fundamental components, HPM allows developers to specify system-level responses to the surrounding environment. Unlike traditional IoT models, which often fall short in scalability and adaptability for SoSs, HPM integrates secure, dynamic coalitions and programmable interactions to handle the complexities of dynaSoSs. This approach ensures that holons can effectively manage interactions, respond to environmental changes, and maintain secure and efficient collaborations within a scalable framework.

\begin{figure}
  \centering
  \includegraphics[width=\columnwidth]{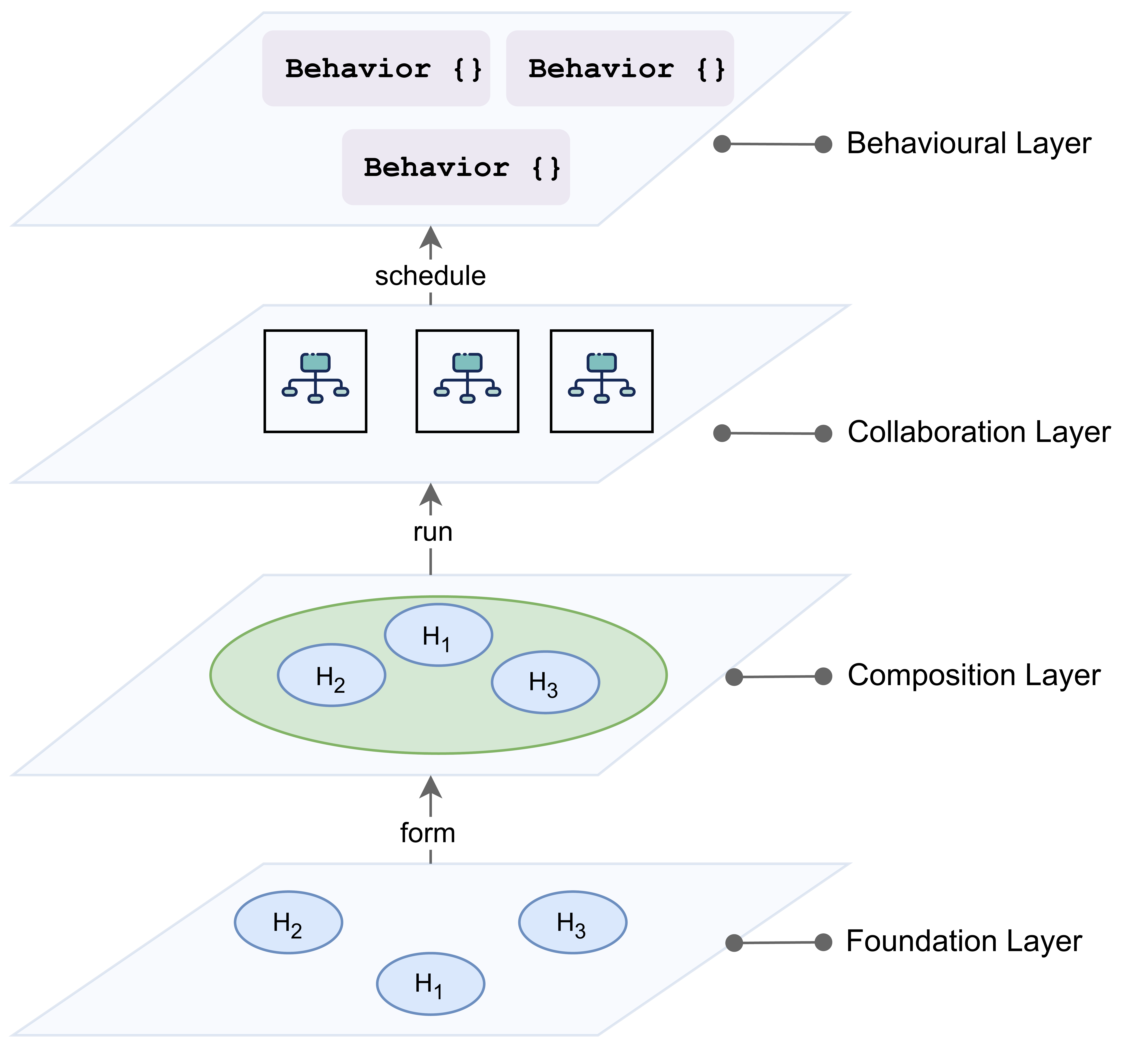}
  \caption{The layered view of the Holon Programming Model (HPM)}
  \label{fig:hpm_layered}
\end{figure}

\begin{figure*}
  \centering
  \includegraphics[width=\textwidth]{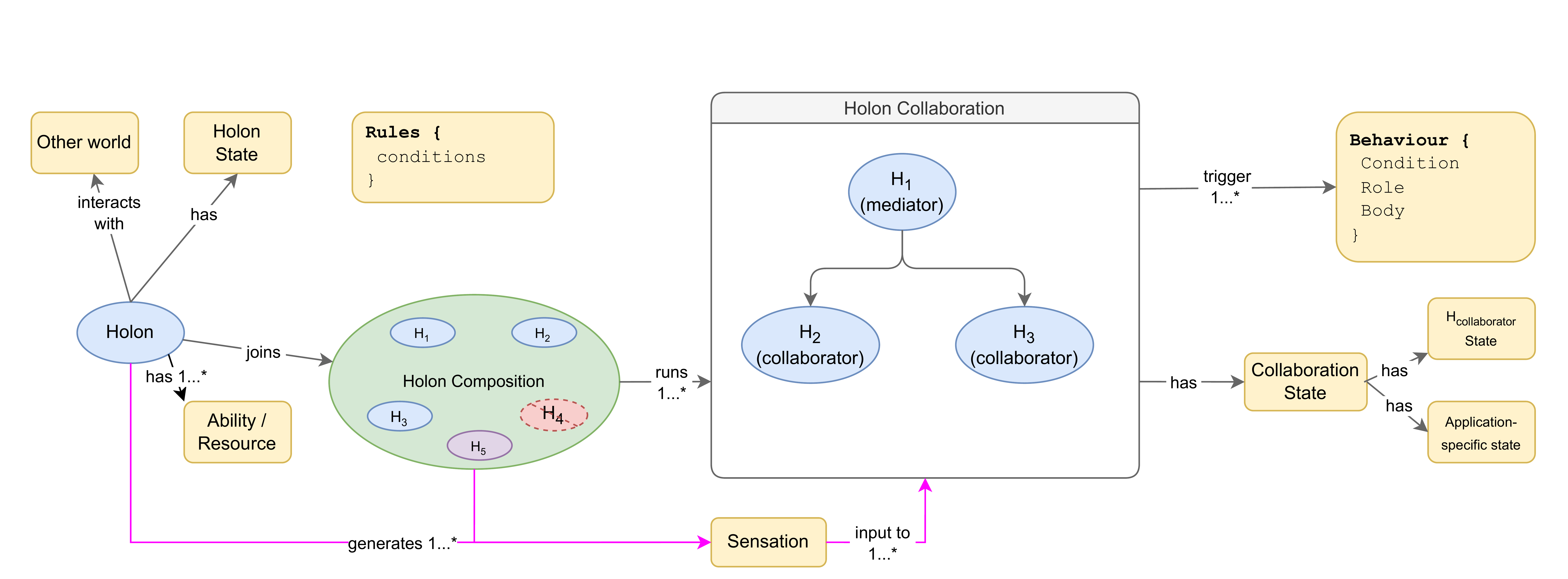}
  \caption{Illustration of the Holon Programming Model (HPM). Each Constituent System (CS) is represented as a holon. The holons make the holon composition with other holons, representing a System of Systems (SoS). The holons in this composition collaborate to trigger behaviours in response to sensations coming from the external environment. }
  \label{fig:HPM}
\end{figure*}

Fig.~\ref{fig:hpm_layered} illustrates the layered view of the HPM, depicting its four distinct layers.
1) The \textit{Foundation Layer} comprises individual holons that encapsulate the CSs.
2) These holons merge to form a \textit{holon composition} at the \textit{Composition Layer}. 
3) The \textit{Collaboration Layer} represents the collaborative efforts of holons to achieve specific objectives.
4) The \textit{Behavioural Layer} outlines the orchestrated behaviours emerging from these collaborations. 
Fig.~\ref{fig:HPM} shows the detailed interior view of the HPM. The remainder of this section provides an in-depth explanation.

\begin{figure*}
  \centering
  \includegraphics[width=\textwidth]{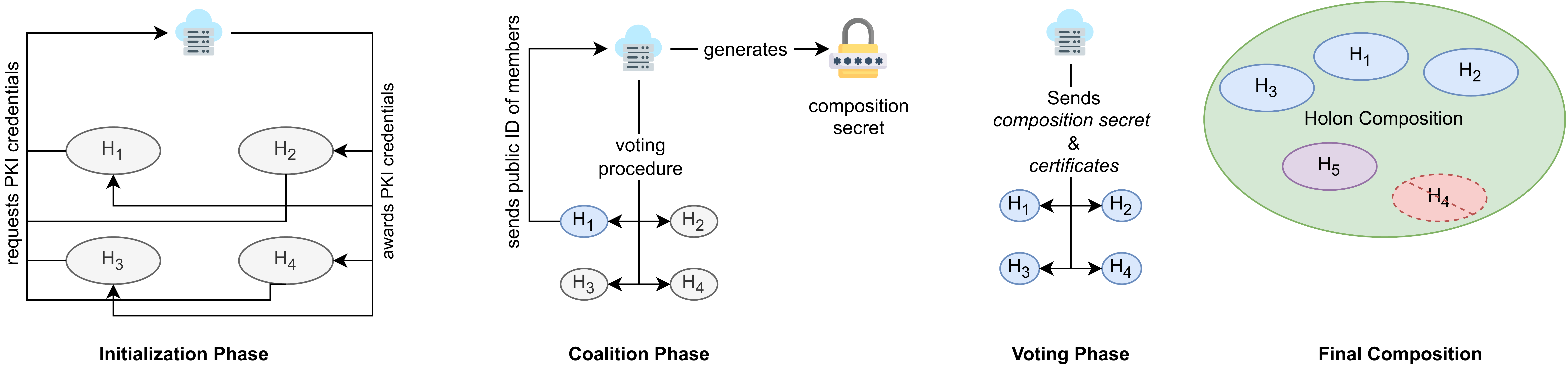}
  \caption{Detailed process of the Holon Composition in the Holon Composition Framework (HCFW).}
  \label{fig:holon_composition}
\end{figure*}

\subsection{Foundation Layer}

This layer consists of holons that encapsulate systems and devices, offering their services or resources as \textit{capabilities}.
For example, a holon might possess a \textit{Natural Language Processing} capabilities, enabling it to translate voice instructions from another holon into its service calls~\cite{icsoft24}.
Additionally, each holon maintains a \textit{state} that provides up-to-date information about the status of its resources and services.

Holons also use their capabilities to interact with and produce output to the \textit{external world}. 
An input from the external digital or physical world is called a \textit{sensation}. 
Holons observe these sensations and broadcast them to the \textit{holon collaboration}.
These observing holons reside in either the \textit{Foundation Layer} or the \textit{Composition Layer}. For example, a holon might generate a sensation to signal its intent to initiate or join a holon composition.

\subsection{Composition Layer}

The core principle of holonic architecture is that the holons can compose to form higher-order holons.
To support this composition, we propose the \textit{Holon Composition Framework} (HCFW), a component of the HPM.

In HCFW, holons compose with each other to create a holon composition in which they collaborate to achieve high-level goals. 
Each holon composition is itself a holon---acting both as a self-sufficient whole in achieving collective objectives and as a part when merging with other compositions.
When two or more holon compositions merge, they form a new composition, wherein the holons and behaviours represent the union of the contributing compositions.

The HCFW conceptually resembles our previously proposed Trusted Device Coalition Framework~\cite{ makitaloActionOrientedProgrammingModel2019}, which enables secure communication among IoT devices.
However, the HCFW extends this concept by supporting system-to-system and system-to-device communication.



In the HCFW, a secure composition of holons is established by engaging holons in trust negotiations. Fig.~\ref{fig:holon_composition} details the formation process of this composition. The process consists of three phases:

\begin{enumerate}
    \item \textit{Initialization Phase}: This phase requires stable connectivity to a trusted entity (e.g., a cloud service). Holons receive certificates from the trusted entity with the corresponding secret and public keys. This requirement ensures direct and secure connectivity with each relevant holon. 
    \item \textit{Coalition Phase}: This phase begins when a holon seeks to compose with other holons. It sends the public identifiers of potential composition members to the trusted entity. The composition secret is then generated and distributed among the composition partners. 
    \item \textit{Voting Phase}: The trusted entity verifies whether the potential partner holons agree to the composition in this phase. Upon confirmation, the trusted entity delivers the composition secret and certificates to the holons. 
\end{enumerate}

This process leads to the forming of the holon composition, which can then operate autonomously. 
The established composition can self-organize by dynamically including or excluding holons through an automated voting mechanism based on predefined rules and the system's assessment of the current situation.
The developer can customize the voting procedure to meet specific requirements, including the percentage of agreement needed for a decision to be implemented across the entire composition.


The newly-formed holon composition needs some mechanisms and processes that the holons must follow to join the composition.
In the HCFW, developers can program such \textit{rules} to specify how the holons should be composed and what \textit{conditions} cause the holon composition to start discovering new holons.
The conditions can be related to any physical or digital world processes using \textit{logical} ($\land$, $\lor$, $\neg$), \textit{temporal} (BEFORE, AFTER, DURING), \textit{arithmetic} ($>$, $<$, $=$, $\neq$), and \textit{aggregation} (COUNT, AVERAGE, SUM) operators. Operands can be any measurable quantity from the physical world (e.g., temperature, proximity) or digital events (e.g., system failure signal, emergency signal). 


While the communication technology used for realizing the HCFW can be any, we recommend choosing protocols that prioritize security, support ad-hoc CS integration and disintegration, and relieve developers from managing composition intricacies in their application logic.

\subsection{Collaboration Layer}

Holon collaborations are an organized subset of holons from the composition. In a collaboration, one holon assumes the role of \textit{mediator} to facilitate interactions, while the others act as \textit{collaborators}, engaging in information exchange. 

Each holon collaboration maintains a  \textit{state} that encompasses the individual states of the participating holons and the collective state of the collaboration.
This shared state is synchronized across all the holons involved in the collaboration. 

The \textit{mediator holon} captures sensations and assembles holons that can fulfil the \textit{roles} required by the \textit{behaviours} associated with these sensations. 
It then executes the code for scheduling the behaviours and manages the collaboration state. 
The mediator holon is selected based on specific criteria, such as superior connectivity, computational power, battery life, diverse capabilities, or other factors the developer determines.
The collaborator holons cooperate with the mediator to implement behaviours.

When two compositions merge, their collaboration layers start sharing the state information. Both compositions dispatch sensations to the selected mediator holon, forwarding them to all collaborators.

Holon collaborations extend the concept of holarchies as they coordinate closely with their environment through a software layer that manages behaviour and scheduling, blending structural integrity with functional agility.

\subsection{Behavioural Layer}

A \textit{behaviour} is the collective response of holons to a sensation triggered by their collaboration. It is a modular unit that determines how holons with collective intelligence interact over a specific period. Each behaviour comprises three key components:

\begin{enumerate}
    \item \textit{Triggering Condition}:  The behaviour is initiated when this condition is met. It also verifies the availability of the necessary holon resources to execute the behaviour.

    \item \textit{Responsibility}: This component specifies which holons are eligible to participate in the behaviour's execution. It maps the required responsibilities to the capabilities of the collaborating holons, ensuring that holons with the necessary and available capabilities are selected for the task.

    \item \textit{Body}:  This part contains the programming logic defined by the developer, outlining the cooperative interactions among holons while leveraging their capabilities.
\end{enumerate}

Behaviours are typically transient, with holon capabilities utilized exclusively during the execution phase. If a holon's capability is unavailable, the behaviour's execution is halted and rescheduled once the capability becomes available.

\section{Scenario: Disaster Management System}
\label{sec:scenario}

To demonstrate the potential utility of the HPM, we present a scenario within a disaster management system. 
This scenario illustrates how the HPM can be applied to enable various CSs within an SoS to respond effectively to environmental stimuli and achieve their objectives through coordinated collaboration and the execution of programmed behaviours.

Disaster management involves a coordinated response to crises, involving multiple entities with the necessary expertise and resources. 
These entities operate heterogeneous cyber-physical systems crucial for maintaining situational awareness and enabling coordination among organizations and residents during disaster scenarios~\cite{fan2018establishing}. However, this process presents several challenges, including effective communication, coordination, resource allocation, and timely information management~\cite{khanSystematicReviewDisaster2023}.

\begin{figure*}
  \centering
  \includegraphics[width=.9\textwidth]{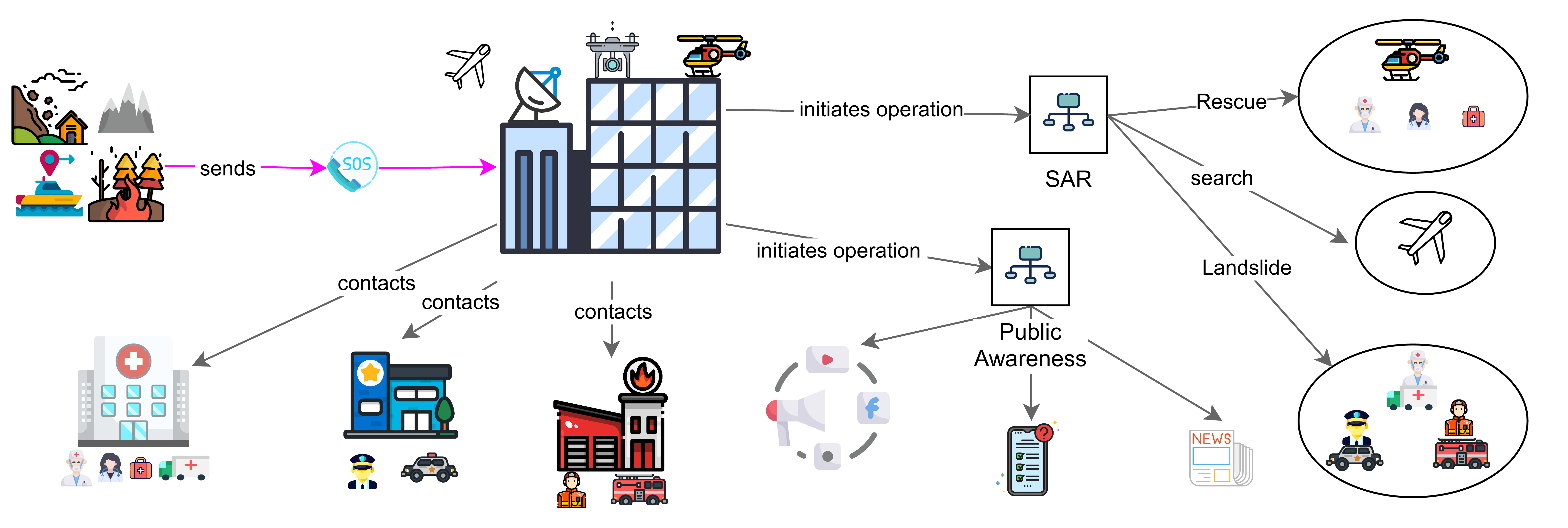}
  \caption{Overview of the Disaster Management System.}
  \label{fig:disaster_management_system}
\end{figure*}

Fig.~\ref{fig:disaster_management_system} presents a disaster management scenario involving various entities and their assets, including Command and Control (C2), government departments, and procedures. In this scenario, an SOS\footnote{Please notice the difference with \textit{SoS}---the acronym for System of Systems} call is initiated from a disaster site. C2 receives the call and immediately begins gathering more detailed information, including determining the nature of the disaster (e.g., earthquake, wildfire, stranded or missing person) and pinpointing its location (a last known position, exact coordinates, or an approximate area). This information helps C2 assess the situation and plan an appropriate response.
C2 has various assets to facilitate the response, including a search plane, a rescue helicopter, and a Micro Aerial Vehicle (MAV). Simultaneously, C2 communicates with other departments, including fire, police, and health services. Based on the assessment, C2 initiates appropriate operations. The figure illustrates two operations: Public Awareness and Search and Rescue (SAR).

\begin{figure*}
  \centering
  \includegraphics[width=.9\textwidth]{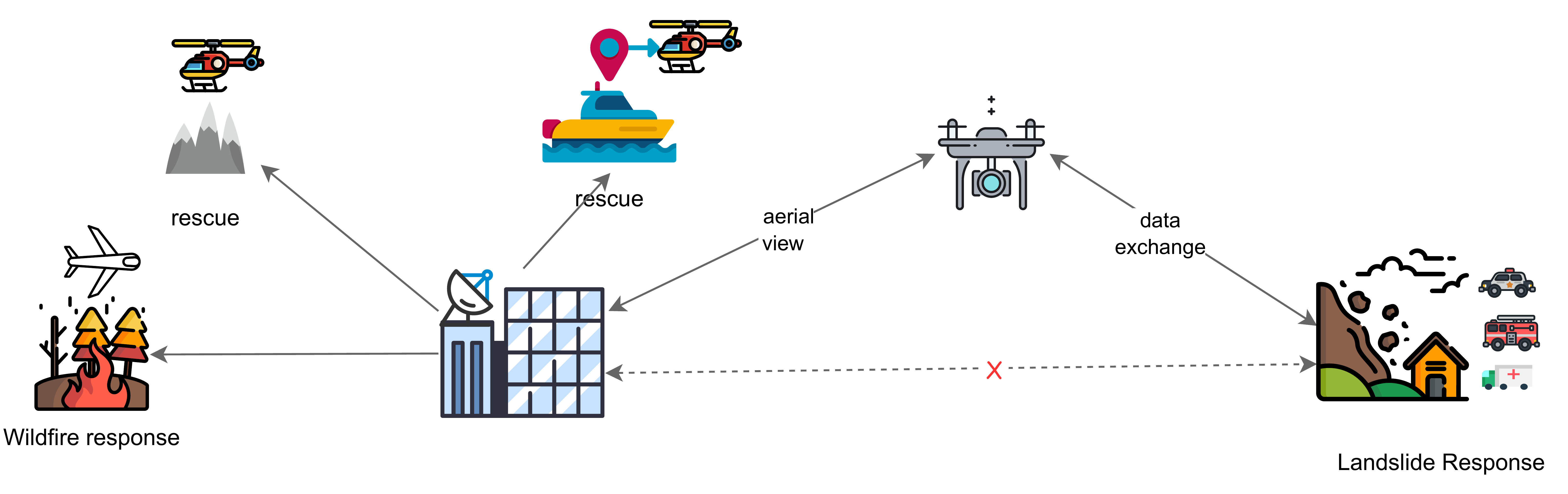}
  \caption{The details of Search and Rescue (SAR) Operation.}
  \label{fig:SAR_operation}
\end{figure*}

Fig.~\ref{fig:SAR_operation} elaborates on the SAR operation, showing how C2 assembles and deploys response teams composed of the necessary personnel and assets. If the SOS call concerns a missing person, C2 deploys a search plane to locate the individual. Once the search is successful, C2 determines whether the person is stranded (e.g., in the ocean or mountains) and, if necessary, dispatches a rescue helicopter equipped with a medical team. For wildfire incidents, C2 mobilizes a search plane outfitted with water-spraying equipment.

In the case of a landslide, C2 must coordinate with other departments due to the complexity of the response, which requires a broader range of capabilities. The landslide site demands paramedic staff and ambulances to provide first aid and transport critically injured individuals to the hospital. 
A fire engine is also required due to the fire risk from potential short circuits. 
Additionally, police presence is essential to safeguard the property of the affected individuals. 
C2 coordinates with the relevant departments to mobilize these resources.

Upon arrival at the disaster site, the assets begin their activities by forming a cohesive team. Throughout the operation, the team maintains constant communication with C2, providing updates on casualties, injuries, and the operation status, while also receiving further instructions. However, due to the remote location, the communication link between the SAR team and C2 is prone to disruptions. C2 deploys an MAV as a communication relay to address this, ensuring a stable connection and seamless coordination throughout the operation.

For the public awareness operation, C2 disseminates information about the disaster through official communication channels, including social media and local news outlets.

\subsection{SoS Perspective and Challenges}

This disaster management system is fundamentally an SoS that demands high levels of adaptability and interoperability. Each entity within this system, such as C2 and various departments, functions as a CS in its own right. 
These heterogeneous CSs must dynamically compose into a cohesive SoS at runtime to execute specific operations, including SAR and Public Awareness. This dynamically composed SoS needs to respond adaptively to environmental inputs by triggering appropriate behaviours essential for efficient operation within the environment.

Developing software for such an SoS presents numerous challenges that traditional programming paradigms often struggle to address adequately. These challenges include ensuring seamless cooperation among C2, rescue teams, and MAVs under unreliable network conditions and executing various SAR operation tasks, such as team deployment, search and rescue, and ad-hoc bridging of MAVs. To meet these challenges, software developers must design behaviours that enable the dynamic composition of teams, allowing the SoS to harness the diverse capabilities of its CSs and respond adaptively to environmental stimuli and operational demands in the disaster management system.

\subsection{Realization of the scenario in HPM}
\begin{figure*}
  \centering
  \includegraphics[width=.9\textwidth]{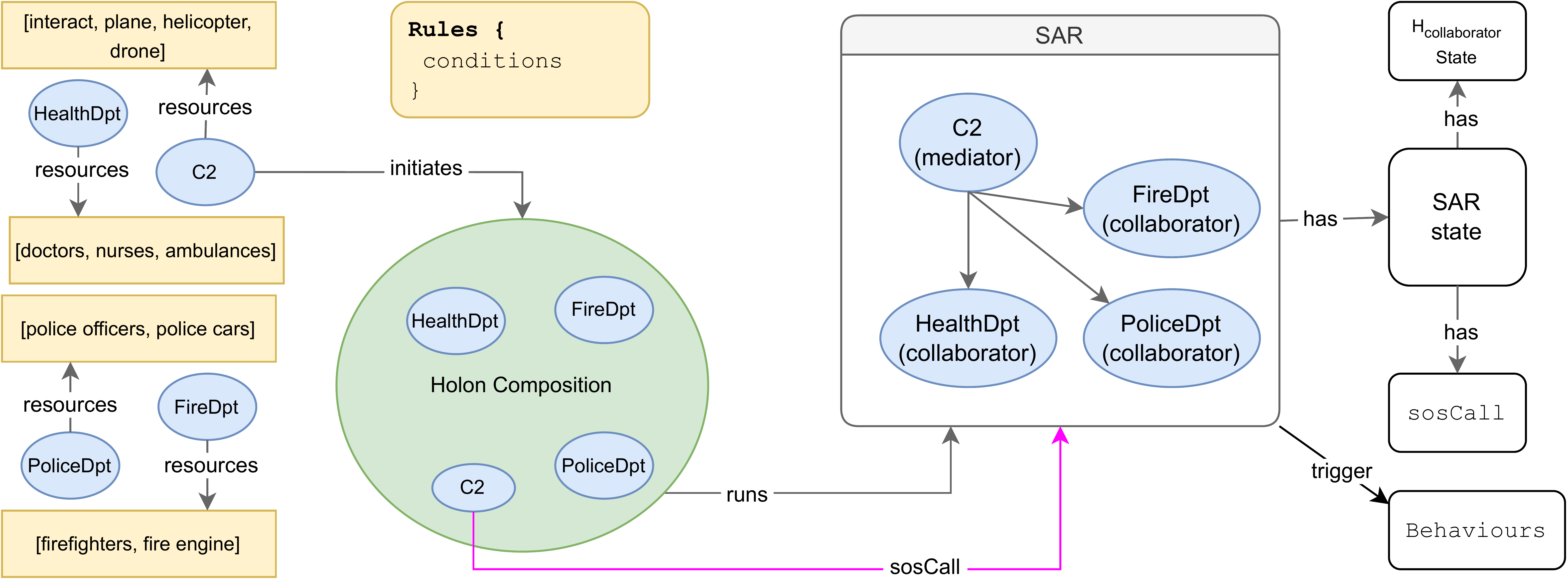}
  \caption{The disaster management system, modelled in the Holon Programming Model (HPM).}
  \label{fig:disater_management_HPM}
\end{figure*}

This section demonstrates how the HPM enables software-defined behaviours and adaptations in this SoS. Fig.~\ref{fig:disater_management_HPM} illustrates entities such as \textit{C2}, \textit{HealthDpt}, \textit{PoliceDpt}, and \textit{FireDpt} as holons, each equipped with specific resources. 
The C2 holon’s resources include a search plane, a rescue helicopter, and an MAV. 
The FireDpt holon has firefighters and fire engines, the PoliceDpt holon is equipped with police officers and police cars, and the HealthDpt holon has doctors, nurses, and ambulances.

The C2 holon interacts with the external world, from which it receives an initial SOS sensation. Upon receiving this sensation, C2 begins gathering more detailed information about the disaster. 
After gathering the information, C2 evaluates the content and initiates the formation of a composition that includes HealthDpt, PoliceDpt, and FireDpt holons. This composition then collaborates to execute specific behaviours in response to the disaster.
The Fig.~\ref{fig:disater_management_HPM} represents the SAR collaboration, where C2 is the mediator while the other departments are collaborators.

Fig.~\ref{fig:SAR_Collab} provides the pseudocode for the SAR collaboration. The collaboration maintains a state object \texttt{sosCall}, initially set to \texttt{null}. This object is populated once C2 detects an SOS sensation, assigning the sensation's location to \texttt{sosCall.loc} and the disaster nature to \texttt{sosCall.type}. The disaster's nature could be a landslide, wildfire, stranded individual, or missing person. The collaboration triggers several behaviours based on \texttt{sosCall.type}.

The \texttt{wildfireResp} behaviour is triggered by the \textit{condition} that  \texttt{sosCall.type} is \texttt{"wildfire"} and the \texttt{waterCarrier} role is fulfilled by a resource from a holon participating in the collaboration. 
Additionally, the resource must have a full water tank and sprayer. 
C2 Holon's search plane is preferred for this role over the helicopter due to its lower fuel consumption, longer flight times, and lack of need for landing at the site.
The \texttt{waterCarrier} proceeds to \texttt{sosCall.loc}, sprays water, and returns to base either upon depleting its tank or upon receiving a command from C2.

The \texttt{search} behaviour is triggered if the condition \texttt{sosCall.type == "missing person"} is met. This behaviour needs a \texttt{searcher}, which C2 holon's search plane can fulfil. Operationally, the \texttt{searcher} moves to the \texttt{sosCall.loc} and conducts the search. Upon completing the search, it returns to the base and sets the \texttt{sosCall.type} to \texttt{"rescue"}, thereby triggering the rescue behaviour.

The \texttt{rescue} behaviour is triggered if \texttt{sosCall.type} is \texttt{"stranded"}. In addition to the rescue helicopter, this behaviour requires the role of paramedic staff, which can be fulfilled by the resources of the HealthDpt holon. The helicopter transports the paramedic staff to the site, where they provide first aid if necessary, retrieve the victim, and then return to base.

The \texttt{landslideResp} behaviour is triggered when \texttt{sosCall.type} is \texttt{"landslide"}. All the holons in the composition (e.g., C2, HealthDpt,  FireDpt, PoliceDpt) are needed to fulfil the necessary roles. 
C2 alerts all entities and ensures their teams are dispatched to the site to perform their specific duties.
The resources from FireDpt, HealthDpt, and PoliceDpt are crucial for mitigating fire hazards, providing medical attention, and securing the property of the affected individuals.
C2 coordinates with these teams and facilitates data exchange.
If the communication link between C2 and the on-site teams weakens, C2 deploys an MAV to establish a communication bridge. This MAV enhances C2's communication capabilities and situational awareness by live-streaming video from the disaster site.

\begin{figure}
\begin{lstlisting}[
    language=Python, 
    numbers=none, 
    columns=fullflexible,
    breaklines=true,
    postbreak=\mbox{\textcolor{red}{$\hookrightarrow$}\space},
    xleftmargin=15pt,
    xrightmargin=15pt,
    frame=single,
    ]
Holon Collaboration SAR {
sos: sosCall := null
  
behaviour wildfireResp(waterCarrier):
  if sosCall.type is "wildfire" and
     waterCarrier has waterTank and
     waterCarrier has waterSprinkler and
     waterCarrier.waterLevel is full:   
  do
    move(waterCarrier, sosCall.loc)
    sprinkleWater(waterCarrier)
    if waterCarrier.tank is empty or
       C2.sensation is "comeback":
      returnToBase(waterCarrier)

behaviour search(searcher):
  if sosCall.type is "missing person":
  do
    move(searcher, sosCall.loc)
    search()
    if search.successful:
      returnToBase(searcher)
      set sosCall.type == "rescue"

behaviour rescue(helicopter, medic):
  if sosCall.type in ["rescue", "stranded"]:
    move(helicopter, sosCall.loc)
    # provide first aid if needed
    returnToBase(helicopter)

behaviour landslideResp(entities[]):
  if sosCall is "landslide":
    async for e in entities:
      alert(e)
      await e.status == "on_site"
      coordinate(C2, e)
    if commLink is "weak":
      deployMAV(sosCall.loc)
  
}
\end{lstlisting}
\caption{The pseudocode of behaviours for Search and Rescue (SAR) collaboration.}
\label{fig:SAR_Collab}
\end{figure}

\section{Conclusion}
\label{sec:conclusion}
This paper introduced the HPM, a novel conceptual approach for programming SoS-level behaviours. The HPM integrates SDS principles with holonic architecture to address critical challenges in SoS programming, including dynamic composition, interoperability, and adaptive behaviour. The HPM enables holons to dynamically join or leave compositions based on their availability and situational context by facilitating the emergence of novel situations and the evolutionary development of collaborative behaviours.

We conceptually demonstrated the HPM's potential through a disaster management scenario, showcasing its ability to orchestrate effective responses to environmental stimuli and achieve objectives through coordinated collaboration. This theoretical exploration highlights the model's promise in managing complex interactions within dynamic environments.

Future work will focus on refining the HPM and implementing the proposed scenarios in simulated and real-world environments. We will evaluate HPM to assess the model's performance, adaptability, and practicality in diverse contexts. Additionally, we will investigate the HPM's performance implications, enhance its security features, and develop supportive tools to facilitate its wider adoption across various SoS domains.

By progressing from conceptualization to practical application, we aim to validate the HPM as a significant step toward more effective SoS programming, ultimately paving the way for creating adaptive, resilient, and scalable systems in increasingly complex and dynamic environments.

\bibliographystyle{IEEEtran}    
\bibliography{main}

\end{document}